\documentclass{article}
\usepackage{ijcai20}

\usepackage{graphicx}
\usepackage{pdfpages}

\usepackage{times}
\usepackage{soul}
\usepackage[utf8]{inputenc}
\usepackage[small]{caption}

\usepackage{amsmath}
\usepackage{amsthm}
\usepackage[ruled,linesnumbered,vlined]{algorithm2e}

\usepackage{url}
\usepackage[hidelinks]{hyperref}
\usepackage{tabularx}
\usepackage{booktabs}
\usepackage{makecell}
\usepackage{diagbox}
\usepackage{multirow}
\usepackage{caption}
\usepackage{subcaption}
\newcolumntype{b}{X}
\newcolumntype{s}{>{\hsize=.5\hsize}X}

\makeatletter
\def\adl@drawiv#1#2#3{%
        \hskip.5\tabcolsep
        \xleaders#3{#2.5\@tempdimb #1{1}#2.5\@tempdimb}%
                #2\z@ plus1fil minus1fil\relax
        \hskip.5\tabcolsep}
\newcommand{\cdashlinelr}[1]{%
  \noalign{\vskip\aboverulesep
           \global\let\@dashdrawstore\adl@draw
           \global\let\adl@draw\adl@drawiv}
  \cdashline{#1}
  \noalign{\global\let\adl@draw\@dashdrawstore
           \vskip\belowrulesep}}
\makeatother

\usepackage{natbib}

\newcommand{\fn}[1]{\footnote{#1}}

\usepackage{CJKutf8}

\usepackage{comment}

\usepackage{verbatim}

\usepackage{colortbl}
\usepackage{xcolor}
\definecolor{cR}{RGB}{255,204,204}
\definecolor{cG}{RGB}{204,255,204}
\definecolor{cB}{RGB}{204,204,255}
\definecolor{cK}{RGB}{221,221,221}
\definecolor{navyblue}{rgb}{0.0,0.0,0.5}

\usepackage[hidelinks]{hyperref} 
\hypersetup
{
    colorlinks=true,
    citecolor=navyblue,
    linkcolor=navyblue,
    filecolor=navyblue,
    urlcolor=navyblue,
}
\newcommand{\Fig}[1]{Figure~\ref{fig:#1}}
\newcommand{\Tab}[1]{Table~\ref{tab:#1}}
\newcommand{\Algo}[1]{Algorithm~\ref{algo:#1}}

\newcommand{\ra}{{$\rightarrow$}}

\title{Pre-training via Leveraging Assisting Languages and Data Selection \\for Neural Machine Translation}

\author
{
Haiyue Song$^1$\hspace{1em}\and
Raj Dabre$^2$\hspace{1em}\and
Fei Cheng$^1$\hspace{1em}\and
Zhuoyuan Mao$^1$\hspace{1em}\and
Sadao Kurohashi$^1$\hspace{1em}\And
Eiichiro Sumita$^2$\\
\affiliations
$^1$Kyoto University\\
$^2$National Institute of Information and Communications Technology\\
\emails
 \texttt{\{song, feicheng, zhuoyuanmao, kuro\}@nlp.ist.i.kyoto-u.ac.jp},\\
\texttt{\{raj.dabre, eiichiro.sumita\}@nict.go.jp}
}

\begin{document}

\maketitle

\begin{abstract}
Sequence-to-sequence (S2S) pre-training using large monolingual data is known to improve performance for various S2S NLP tasks in low-resource settings. However, large monolingual corpora might not always be available for the languages of interest (LOI). To this end, we propose to exploit monolingual corpora of other languages to complement the scarcity of monolingual corpora for the LOI. A case study of low-resource Japanese--English neural machine translation (NMT) reveals that leveraging large Chinese and French monolingual corpora can help overcome the shortage of Japanese and English monolingual corpora, respectively, for S2S pre-training. We further show how to utilize script mapping (Chinese to Japanese) to increase the similarity between the two monolingual corpora leading to further improvements in translation quality. Additionally, we propose simple data-selection techniques to be used prior to pre-training that significantly impact the quality of S2S pre-training. An empirical comparison of our proposed methods reveals that leveraging assisting language monolingual corpora, data selection and script mapping are extremely important for NMT pre-training in low-resource scenarios.
\end{abstract}

\section{Introduction}
Neural Machine Translation (NMT) \citep{DBLP:journals/corr/SutskeverVL14,DBLP:journals/corr/BahdanauCB14:original} is known to give state-of-the-art (SOTA) translations for language pairs with an abundance of parallel corpora. However, most language pairs are resource poor (Russian--Japanese, Marathi--English) as they lack large parallel corpora and it is possible to compensate the lack of bilingual training data by leveraging large monolingual corpora. One popular approach for this is data augmentation,  e.g. by back-translation \citep{sennrich-etal-2016-improving} of monolingual data to produce pseudo-parallel corpora. The limitation of this approach is that the initial MT systems used for back-translation should be robust enough to yield pseduo-parallel corpora of decent quality. Recently, another approach has gained popularity where the NMT model is pre-trained through tasks that only require monolingual data \citep{song2019mass, qi-etal-2018-pre}. 

Pre-training has seen a surge in popularity in NLP ever since models such as BERT \citep{DBLP:journals/corr/abs-1810-04805} have led to new state-of-the-art results in text understanding. However, BERT-like models were not designed to be used for NMT  in the sense that they are essentially language models and not sequence to sequence (S2S) models. To address this, \citet{song2019mass} recently proposed MASS, a S2S specific pre-training task for NMT and obtained new state-of-the-art results in low-resource settings. MASS assumes that a large amount of monolingual data is available for the languages involved but this may not always be true. Such language pairs that lack both parallel corpora and monolingual corpora are ``truly low-resource'' and challenging. 

Fortunately, languages are not isolated and often belong to ``language families'' where they have similar orthography (written script; shared cognates) or similar grammar or both. \citet{guzman-etal-2019-flores} leveraged linguistically similar resource-rich Hindi to improve translation involving Nepalese through unsupervised NMT and semi-supervised NMT. Motivated by this, in this paper we hypothesize that we should be able to leverage large monolingual corpora of other (assisting) languages to help the monolingual pre-training of NMT models for the languages of interest (LOI) that may lack monolingual corpora. Methods such as MASS focus more on the pre-training task but not as much on the data being used. In the past, data pre-processing and filtering methods shown to impact MT performance as they minimize the ``differences'' between the datasets used for training MT systems \citep{axelrod-etal-2011-domain}.  Following this, we further hypothesize that subjecting the pre-training corpora to script mapping and simple but rigorous data-selection techniques should help minimize the vocabulary and distribution differences, respectively, between the pre-training, main training (fine-tuning) and testing time datasets. This should help the already consistent pre-training and fine-tuning objectives leverage the data much better and thereby, possibly, boost translation quality.

To this end, we experiment with ASPEC Japanese--English translation in a variety of low-resource settings for the parallel corpora. We focus on sequence to sequence pre-training using MASS that uses a variety of sizes of Japanese and English monolingual corpora filtered with data-selection techniques to establish strong baselines. We then simulate low monolingual corpora situations for Japanese and English and complement the lack of corpora using Chinese (for Japanese) and French (for English). Our experiments reveal that while it's possible to leverage \textbf{unrelated languages} for pre-training, using \textbf{related languages} is extremely important. We identified that, it is absolutely important to maximize the similarities between the assisting languages and the languages of interest. We showed that Chinese to Japanese script mapping can boost translation quality by maximum 7.4 B\caps{leu} in a realistic low-parallel and monolingual corpus setting. 

The contributions of our work are as follows:

\noindent\textbf{1. Leveraging assisting languages:} We give a novel study of leveraging monolingual corpora of related and unrelated languages for NMT pre-training.\\
\noindent\textbf{2. Pre-processing:} We show the strong impact of robust and rigorous data-selection and script mapping on the translation quality.\\
\noindent\textbf{3. Empirical evaluation:} We make a comparison of existing and proposed techniques in a variety of corpora settings to verify our hypotheses.\\
\noindent\textbf{4. Analysis:} We analyze our results and share the lessons we have learned.


\section{Related work}
Our research is at the intersection of works on monolingual pre-training for NMT and leveraging multilingualism for low-resource language translation. 

In monolingual pre-training\footnote{This is an instance of ``transfer learning" just like Cross-lingual transfer. ``Pre-training" often implies that the training task differs from the target task.} approaches, all or part of a model is first trained on tasks that require monolingual data.  Pre-training has enjoyed great success in other NLP tasks with the development of methods like BERT \citep{DBLP:journals/corr/abs-1810-04805}.
To address this, \citet{song2019mass} recently proposed MASS, a new state-of-the-art NMT pre-training task that jointly trains the encoder and the decoder. Our approach builds on the initial idea of MASS, but focuses on complementing the potential scarcity of monolingual corpora for the languages of interest using relatively larger monolingual corpora of other (assisting) languages.

On the other hand, leveraging multilingualism involves cross-lingual transfer which solves the low-resource issue by using data from different language pairs. One can use a richer language pair \citep{DBLP:conf/emnlp/ZophYMK16:original}, or several language pairs at once \citep{dabre-etal-2019-exploiting,dong-etal-2015-multi}.  \citet{rudramurthy19}  also proposed to re-order the assisting languages to be similar to the low-resource language. \citet{dabre-etal-2017-empirical} showed the importance of transfer learning between languages belonging to the same language family but corpora might not always be available in a related language. A mapping between Chinese and Japanese characters \citep{chu-etal-2012-chinese} was shown to be useful for Chinese--Japanese 
dictionary construction \citep{dabre-etal-2015-large}.
Such mappings between scripts or unification of scripts \citep{hermjakob-etal-2018-box} can artificially increase the similarity between languages which motivates part of our work.

\section{Our Method: Using Assisting Languages}
\begin{figure}[t]
    \centering
    \includegraphics[width=1.0\linewidth]{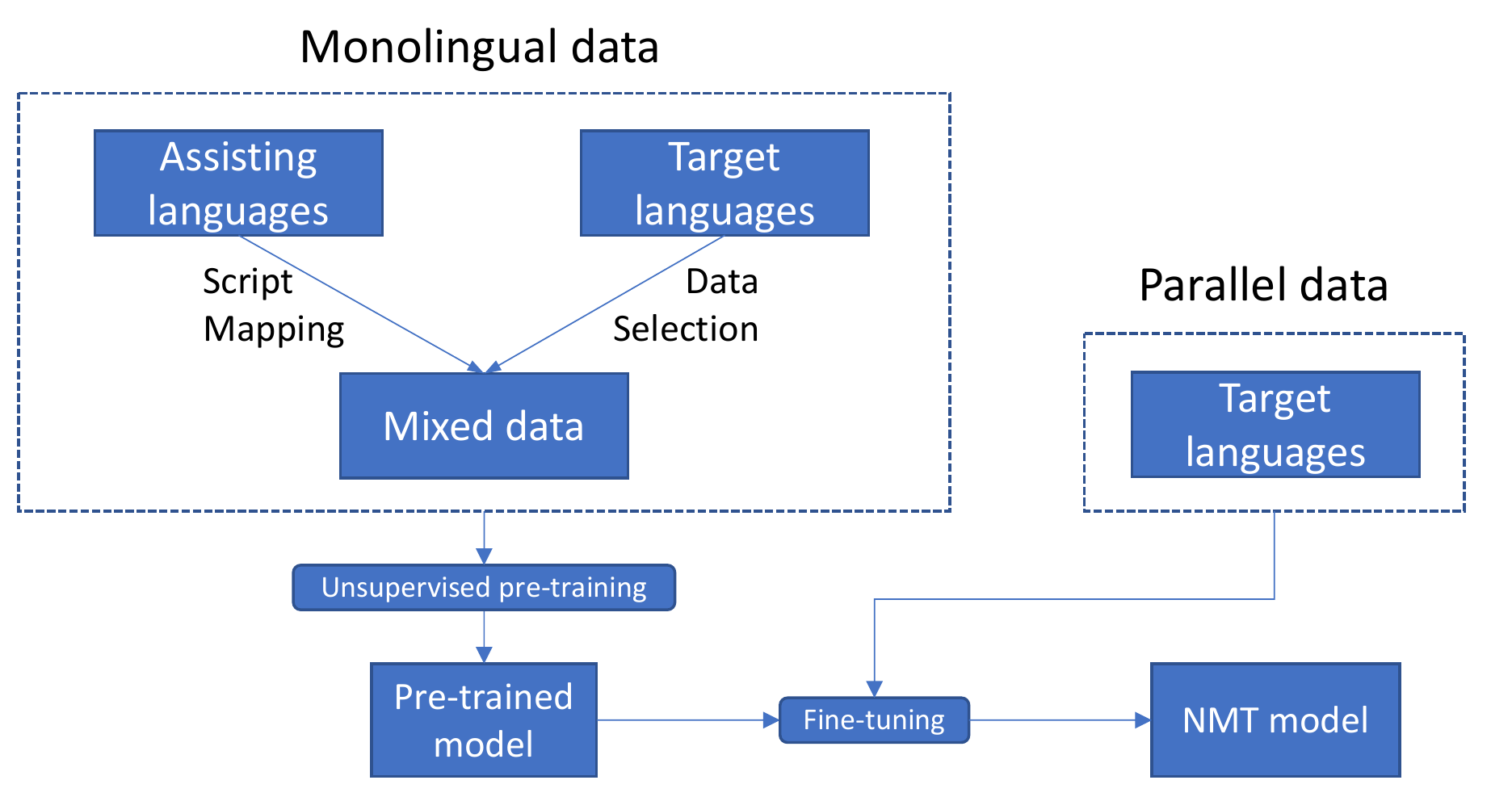}
    \caption{An overview of our proposed method consisting of script mapping, data selection, pre-training and fine-tuning}
    \label{fig:framework}
\end{figure}
We propose a novel monolingual pre-training method for NMT which leverages monolingual corpora of assisting languages to overcome the scarcity of monolingual and parallel corpora of the languages of interest (LOI). The framework of our approach is shown in \Fig{framework} which consists of script mapping, data selection, pre-training and fine-tuning.

\subsection{Data Pre-processing}
Simply pre-training a NMT model on vast amounts of monolingual data belonging to the assisting languages and LOI can improve translation quality. However, divergences between the languages and also the distributions of data between different training phases is known to impact the final result. Choosing linguistically related assisting languages is the best solution. However, even if the assisting languages are linguistically related, a difference in orthography can have a negative impact.  Motivated by past works on orthography mapping/unification \citep{hermjakob-etal-2018-box,chu-etal-2012-chinese} and data selection for MT \citep{axelrod-etal-2011-domain}, we propose to improve the efficacy of pre-training by reducing data and language divergence.

\subsubsection{Script Mapping}
Previous research has shown that enforcing shared orthography \citep{sennrich-etal-2016-neural,dabre-etal-2015-large} has a strong positive impact on translation. Following this, we propose to leverage existing script mapping rules\footnote{Transliteration is another option but transliteration systems are relatively unreliable compared to handcrafted rule tables.} or script unification mechanisms to, at the very least, maximize the possibility of cognate sharing and thereby bringing the assisting language closer to the LOI. This should strongly impact languages belonging to the same family but written in different scripts.

However, for languages such as Korean, Chinese and Japanese there may exist a many to many mapping between their scripts. As such, incorrect mapping of characters (basic unit of a script) might produce wrong words and reduce cognate sharing. We propose two solutions to address this.

\noindent \textbf{1. One-to-one mapping:} Here we do not care about word level information and map each character in one language to its corresponding character in another language. Here, we just select the first mapping in the mapping list. 

\noindent \textbf{2. Many-to-many mapping with LM scoring:} A more sophisticated solution is where for each tokenized word-level segment in one language we enumerate all possible combinations of mapped characters and use a language model in the other language to select the character combination with the highest score as the result. 

\subsubsection{Note: Chinese--Japanese script mapping}
Japanese language is written in Kanji which was borrowed from China. Over time the written scripts have diverged and the pronunciations are naturally different but there are a significant number of cognates written in both languages. As such pre-training on Chinese should benefit translation involving Japanese. \citet{chu-etal-2012-chinese} created a mapping table between them which can be leveraged to further increase the number of cognates. 

\subsubsection{Data Selection}
Existing NMT pre-training mechanisms relying on monolingual data do not focus on data selection. Often, the pre-training monolingual data and the fine-tuning parallel data belong to different domains \citep{axelrod-etal-2011-domain, wang-neubig-2019-target} have shown that proper data selection can reduce the differences between the natures of data between different training phases. In this paper we propose to use a language model (LM) based and a simple but novel sentence length based data selection mechanisms.

\noindent \textbf{1. LM based data selection:}
To select pre-training data which has similar data distribution with target domain, we use language model trained by in-domain data to sort sentences from the highest LM score and use the top N sentences that are expected to be similar to the in-domain data.

\noindent \textbf{2. Length based data selection:} See \Algo{length-distribution-selection}.
We first calculate the target length distribution (the ratio of all lengths in $TargetFile$) and then we fill the length distribution for $InputFile$ by adding lines where the ratio of the length of that line is lower than the upperbound value. These $SelectNum$ lines contain sentences with length distributions similar to the in-domain dataset thereby increasing the similarity between the datasets.

\begin{algorithm}[t]
\SetKwInOut{Input}{Input}
\SetKwInOut{Output}{Output}
\Input{$\mathit{TargetFile}$, $\mathit{InputFile}$, $\mathit{SelectNum}$}
\Output{$\mathit{SelectedLines}$}
$\mathit{TargetDistribution}\gets\{\}$\;
$\mathit{CurrentDistribution}\gets\{\}$\;
$\mathit{SelectedLines}\gets\{\}$\;
$\mathit{TargetNum}=\mathit{\#\;of\;Lines\;in\;TargetFile}$\;
\ForEach {$\mathit{Line}\in\mathit{TargetFile}$}
{
    $\mathit{TargetDistribution[len(Line)]+=1}$\;
}
\ForEach {$\mathit{Line}\in\mathit{InputFile}$}
{
    \If {$\mathit{CurrentDistribution[len(Line)]/SelectNum}<\mathit{TargetDistribution[len(Line)]/TargetNum}$}
    {
        $\mathit{CurrentDistribution[len(Line)]}+=1$\;
        $\mathit{SelectedLines}\gets\mathit{SelectedLines}\cup\{\mathit{Line}\}$\;
    }
}
\caption{Length Distribution Data Selection}
\label{algo:length-distribution-selection}
\end{algorithm}

\subsection{NMT Modeling}
In order to train a NMT model we first use the pre-processed monolingual data for pre-training and then resume training this model on parallel data to fine-tune for the languages of interest. In our work, we do not make any assumptions about the NMT architecture.

\subsubsection{Pre-training and fine-tuning (MASS)}
MASS is a pre-training method for NMT proposed by \citet{song2019mass}. In MASS pre-training, the input is a sequence of tokens where a part of the sequence is masked and the pre-training objective is to predict the masked fragments using an encoder-decoder model. 
The NMT model is pre-trained with the MASS task, until convergence, jointly for both the source and target languages. Thereafter training is resumed on the parallel corpus, a step known as fine-tuning \citep{DBLP:conf/emnlp/ZophYMK16:original}.  We refer the readers to the original paper by \citet{song2019mass} for further details.


\section{Experimental Settings}
We conducted experiments on Japanese--English (Ja--En) translation in a variety of simulated low-resource settings using the ``similar'' assisting language pairs Chinese (Zh) and French (Fr) and the ``distant'' assisting language pairs Russian (Ru) and Arabic (Ar).

\subsection{Datasets}
For the parallel corpora, we used the official ASPEC Ja--En datasets \citep{nakazawa2016aspec} provided by WAT 2019\footnote{\url{http://lotus.kuee.kyoto-u.ac.jp/WAT/WAT2019/index.html\#task.html}}. The official split consists of 3M, 1790 and 1872 train, dev and test sentences respectively. We sampled parallel corpora from the top 1M sentences for fine-tuning. Out of the remaining 2M sentences, we used the En side of the first 1M and the Ja side of the next 1M sentences as monolingual data for language modeling for data selection. Additionally, we used Common Crawl\fn{\url{http://data.statmt.org/ngrams/}} monolingual corpora for pre-training. To train LMs for data-selection of the assisting languages corpora, we used news commentary datasets \fn{\url{http://data.statmt.org/news-commentary/v14/}}. For the rest of this paper we consider the ASPEC and news-commentary monolingual sentences as in-domain and the rest of the pre-training sentences as out-of-domain.

\subsection{Data Pre-processing}

\subsubsection{Normalization and Initial Filtering}




We applied NFKC normalization to data of all languages. Juman++ \citep{jumanpp1} for Ja tokenization, jieba\fn{\url{https://github.com/fxsjy/jieba}} for Zh tokenization and NLTK\fn{\url{https://www.nltk.org}} tokenization for other languages. We filtered out all sentences from the pre-training data that contain fewer than 3 and equal or more than 80 tokens. Especially for Chinese data, which we found to be noisy, we filtered out sentences containing fewer than 30 percent Chinese words or more than 30 percent English words.
\subsubsection{Script Mapping}

Out of all assisting languages Chinese is the only one that can be mapped to Japanese reliably. In this paper, we focused on converting Chinese to Japanese script to make them more similar by using the mapping table from \cite{chu-etal-2012-chinese} and the mapping approaches mentioned in the previous section. French and English are written using the Roman alphabet and do not need any script manipulation. As for Arabic and Russian, we did not perform script mapping in order to show the impact of using distant languages (script-wise as well as linguistically).

\subsubsection{Data selection}
We performed data selection through both implicit and explicit data distribution selection methods.

\noindent\textbf{1. LM Data Selection}:We used language models trained on the in-domain monolingual data and sort the common crawl monolingual sentences on the LM score (highest to lowest) and selected the top $N$ for pre-training. N varies from 1M to 20M sentences. We used KenLM \citep{heafield-2011-kenlm} to train 5-gram LMs.
\noindent\textbf{2. Length Distribution Data Selection}
We used the ASPEC dev set to obtain an estimate of the test time in-domain sentence length distribution. We found that the length distribution of 2,000 lines and 1M lines randomly selected from ASPEC training set are very similar, with same median and similar stdev. This justified our choice of using the dev set to better approximate test set length distributions.


Following \Algo{length-distribution-selection}, when selecting English lines from Common Crawl English monolingual data, we took English side of ASPEC dev set as $TargetFile$, 100M CommonCrawl English monolingual data as $InputFile$ ans 20M as $SelectNum$. We calculated the number of lines in ASPEC dev set as $TargetNum$ and calculated the number of each length of all lines in ASPEC dev set.
We filled the $CurrentDistribution$ line by line in Common Crawl file if the ratio of the length of current line had been less than the ratio of this length in target length distribution. The Japanese data selection was performed similarly with that of English. We did not apply length distribution selection to data of the assisting languages because we lack an equivalent ASPEC-like development set.

Since there is no conflict between LM data selection and length distribution data selection, we tried all four combinations.


\subsubsection{Dataset mixing for pre-training}
We combined monolingual data of all assisting languages and languages of interest (LOI; Japanese and English). When mixing datasets of different sizes, we always oversampled the smaller datasets to match the size the the largest one. 

We experimented with several scenarios when we have different sizes of monolingual data for Japanese and English namely, 1) resource-rich scenario, where we have a lot of monolingual data 2) resource-poor scenario, where we only have a little monolingual data of target languages and 3) zero-resource scenario, where we do not have any monolingual data of target languages. We assumed that we have abundant monolingual data of the assisting languages.

\subsection{Training and Evaluation Settings}
We used the tensor2tensor framework \citep{vaswani2018} \fn{\url{https://github.com/tensorflow/tensor2tensor}, version \href{https://github.com/tensorflow/tensor2tensor/releases/tag/v1.14.0}{1.14.0}.} with its default ``\emph{transformer\_big}'' setting, such as dropout=0.2, attention dropout=0.1, optimizer=adam with beta1=0.9, beta2=0.997.

We created a shared sub-word vocabulary using Japanese and English data from ASPEC mixing with Japanese, English, Chinese and French data from Common Crawl.  We used 
SentencePiece \citep{kudo-richardson-2018-sentencepiece} and obtained a vocabulary with the size of roughly 64k . We used this vocabulary in all experiments except unrelated language experiment where Arabic and Russian were used instead of Chinese and French data.

For all pre-training models, we saved checkpoints every 1000 steps and for all fine-tuning models, we saved checkpoints every 200 steps. We used early-stopping using approximate-B\caps{leu} as target and stops when no gain after 10,000 steps for pre-training and 2,000 steps for fine-tuning. We fine-tuned different fine-tune settings from the last checkpoint of each pre-trained model.

For decoding we averaged 10 checkpoints of the fine-tuning stage with $\alpha=0.6$ and $beamsize=4$. We used sacreB\caps{leu}\fn{\url{https://github.com/mjpost/sacreBLEU}} to evaluate B\caps{leu} score for all translation evaluation.  

\subsection{Models Evaluated}

\subsubsection{Pre-trained Models}

We separated pre-training settings into different blocks as shown in \Tab{res-low-resource}. Baseline model without fine-tuning is shown as A1. Zero (0M), low (1M) and rich (20M) monolingual-corpus scenarios are shown in parts B, C and D, respectively. Pre-trained models aim to identify the best setting considering pre-training data size, mapping method, data-selection method and related languages. These are labelled from E to H respectively.

\subsubsection{Fine-tuned Models}

We evaluated both Ja\ra En and En\ra Ja models with four parallel dataset size settings, 3K, 10K, 20K and 50K, selected from the previously selected 1M ASPEC parallel sentences.

%
%
%
%
%
%
%
%
%
%
%
%
%
%
%

\section{Results and Analysis}

\begin{table*}[!ht]
    \begin{center}
    \small
    \resizebox{\textwidth}{!}
    {
        \begin{tabular}{c|c|cccc|c c c c|c c c c}
            \toprule
            \multirow{3}{*}{\#} & \multicolumn{5}{c|}{\textbf{Pre-training}}  & \multicolumn{8}{c}{\textbf{Fine-tuning}} \\ 
            &\multirow{2}{*}{Data pre-processing}& \multirow{2}{*}{Zh}&\multirow{2}{*}{Ja}&\multirow{2}{*}{En}&\multirow{2}{*}{Fr}&
            \multicolumn{4}{c|}{En\ra Ja} & \multicolumn{4}{c}{Ja\ra En} \\
            &&&&&&  3K &  10K & 20K & 50K& 3K & 10K & 20K & 50K\\
            \midrule
            A1 &-&-&-&-&- & 2.5 & 6.0  & 14.4 & 22.9  & 1.8 & 4.6 & 10.9 & 19.4 \\
            \midrule
            B1 & w/o Zh\ra Ja mapping&20M&20M&20M&20M  & 0.8 & 0.9 & 1.5 & 2.6 & 0.6 & 1.1 & 1.4 & 2.3 \\
            B2 &1-to-1 Zh\ra Ja mapping&20M&20M&20M&20M & \textbf{7.0} & \textbf{13.4} &\textbf{19.3} & \textbf{25.7} & \textbf{5.9} & \textbf{11.1} & \textbf{15.0} & \textbf{19.8} \\
            B3 &LM-scoring Zh\ra Ja mapping&20M&20M&20M&20M & 6.3 & 12.7 & 18.1 & 24.7 & 5.7 & 10.3 & 13.5 & 18.9 \\
            \midrule
            C1 &random&-&20M&20M&-& 7.6 & 14.7 & 19.5 & 26.1 & 6.7 & 11.8 & 14.8 &19.5\\
            C2 &LM-scoring&-&1M&1M& - & 3.5 & 8.9 & 13.7 & 22 & 3.8 & 8.3 & 11.7 & 16.8 \\
            C3 &LM-scoring&-&5M&5M& -& 5.8 & 12.8 & 18.1 & 25.6 & 5.3 & 11.2 & 14.5 & 19.6 \\
            C4 &LM-scoring&-&10M&10M& - & 4.9 & 11.2 & 16.9 & 24.2 & 5.0 & 9.6 & 12.9 & 17.7 \\
            C5 &LM-scoring&-&15M&15M& - & 6.9 & 14.6 & 19.4 & 26.0 & 6.5 & 11.7 & 15.1 & 19.3 \\

            C6 &LM-scoring&-&20M&20M&- & 4.7 & 11.7 & 16.6 & 23.9 & 4.5 & 9.1 & 12.9 & 18.3 \\
            C7 &LD&-&20M&20M&-& \textbf{9.4} & \textbf{17.2} & \textbf{21.7} & \textbf{27.8} & \textbf{8.2} &\textbf{13.8} & \textbf{17.1} & 20.6 \\
            C8 &LM-scoring + LD&-&20M&20M&-& 7.9 & 14.4 & 19.7 & 26.4 & 7.1 & 11.5 & 15.2 & \textbf{20.7} 
      
            \\
            \midrule
            D1 &1-to-1 Zh\ra Ja mapping + LD&20M&-&-&- & \textbf{5.3} & \textbf{14.5} & \textbf{20.0} & \textbf{26.1} & \textbf{3.7} & \textbf{11.2} & \textbf{15.6} & \textbf{20.5} \\
            D2 &LD&-&-&-& 20M & 3.4 & 9.1 & 14.9 & 23.4 & 2.1 & 6.3 & 11.3 & 17.7 \\
            D3 &1-to-1 Zh\ra Ja mapping + LD&20M&-&-& 20M & 2.1 & 6.7 & 12.6 & 21.9 & 2.2 & 6.3 & 10.7 & 16.8 \\
            \midrule
            E1 &LD&-&1M&1M& - & 7.7 & 15.8 & \textbf{20.7} & 26.3 & 7.2 & \textbf{12.7} & 15.7 & 19.6\\
            E2 &1-to-1 Zh\ra Ja mapping + LD&20M&1M&1M& - & \textbf{8.3} & \textbf{16.4} & 20.2 & \textbf{26.9} & \textbf{7.5} & 12.5 & \textbf{16.3} & \textbf{20.7}\\
            E3 &LD&-&1M&1M& 20M & \textbf{8.3} & 15.3 & 19.3 & 26.7 & 6.8 & 12.3 & 15.4 & 20.4\\
            E4 &1-to-1 Zh\ra Ja mapping + LD&20M&1M&1M& 20M & 7.1 & 15.2 & 19.4 & 26.5 & 6.6 & 12.0 & 15.4 & 19.9\\
            \midrule
            F1 &LD&-&15M&15M&- & 9.6 & \textbf{17.2} & 21.5 & \textbf{28.0} & \textbf{8.6} & \textbf{13.5} & \textbf{16.8} & \textbf{20.9} \\
            F2 &1-to-1 Zh\ra Ja mapping + LD&20M&15M&15M& - & \textbf{9.7} & 17.1 & \textbf{21.6} & 27.2 & 8.3 & 13.3 & 16.7 & 20.6 \\
            F3 &LD&-&15M&15M& 20M & 7.7 & 15.0 & 19.8 & 26.3 & 6.3 & 11.7 & 15.1 & 20.2 \\
            F4 &1-to-1 Zh\ra Ja mapping + LD&20M&15M&15M& 20M & 7.7 & 14.9 & 19.7 & 26.1 & 6.5 & 11.4 & 15.4 & 19.8 \\
            \midrule
            G1 &LM-scoring&-&20M&20M& -& 4.7 & 11.7 & 16.6 & 23.9 & 4.5 & 9.1 & 12.9 & 18.3 \\
            G2 &1-to-1 Zh\ra Ja mapping + LM-scoring&20M&20M&20M&20M & \textbf{7.0} & \textbf{13.4} & \textbf{19.3} & \textbf{25.7} & \textbf{5.9} & \textbf{11.1} & \textbf{15.0} & \textbf{19.8} \\
            G3 &LM-scoring + Ar20M + Ru20M&-&20M&20M&-& 4.8 & 12.1 & 18.1 & 25.1 & 4.4 & 10.2 & 13.5 & 18.9 \\
            \bottomrule
        \end{tabular}
    }
    \caption{Low-resource pre-training experiments. LD is with the meaning of ``length distribution''. Best results of each part are in bold.}
    \label{tab:res-low-resource}
    \end{center}
\end{table*}




In \Tab{res-low-resource}, we show results of several experimental settings to analyse the effect of: pre-training data size, Zh\ra Ja mapping methods, data selection methods and choices of unrelated languages versus related languages.

\subsubsection{Chinese to Japanese mapping}

In part B, we compared two mapping methods with the method without mapping at all. Results showed that the one-to-one mapping (character-level mapping) gives better B\caps{leu} score than word-level mapping consistently on most fine-tuning settings, about 0.7 to 1.0 in most cases.

Also, if we do not perform Zh\ra Ja mapping, we found there are a lot of OOVs in Chinese monolingual data which will dramatically affect the model in a negative way leading to very low performance after fine-tuning. This shows the importance behind data consistency between the pre-training and fine-tuning stages.

\subsubsection{Pre-training size and data selection experiments}
We tried all combinations of proposed two data selection strategies for monolingual data and show results in part C of the table.\\
\noindent\textbf{1.} Random selection\\
\noindent\textbf{2.} Using length distribution selection\\
\noindent\textbf{3.} LM score sorting based selection\\
\noindent\textbf{4.} Combining 2 and 3.

Comparing using `random' (C1) and `LM-scoring' (C6) methods, we found that randomly selecting sentences will give better results than using sentences with best LM score. We suppose this is due to the divergence between selected data distribution and true data distribution. The LM may cause a data distribution preference, which is also a kind of bias. We also found that adding an explicit constraint, the length distribution, gave better results. Random+length distribution selection (C7) gave about 2 to 3 B\caps{leu} score improvement compared with random selection (C1). LM score+length distribution (C8) also gave 2 to 3 B\caps{leu} score improvement on most settings over only LM-scoring (C6). 

In part C, we also explored how different sizes of pre-training data affect the translation quality. We saw that using even 1M monolingual Japanese and English data (C2) gives much better results than without pre-training (A1) when fine-tune on small dataset. For example on 10K fine-tuning dataset. We observed about 3 B\caps{leu} score improvement on En\ra Ja side and 3.7 score improvement on Ja\ra En side.

As we added monolingual data gradually, we saw that the B\caps{leu} score improved gradually, where using 5M monolingual data (C3) gave much higher scores than using 1M (C2).
But the B\caps{leu} score is not perfectly consistent with monolingual data size as 15M monolingual data (C5) gives the best results on most of the fine-tuning settings. We did not explore whether this phenomenon is caused by the quality of the monolingual dataset or other factors. We experimented with LM-selection for these data size variation experiments as our purpose was to show that even small amounts of pre-training data is enough to give improvements via fine-tuning. In the future we will explore data size variation for all other data selection approaches.
For the remaining experiments we decided to use length distribution (LD) based data selection and 1 to 1 script mapping wherever applicable.

\subsubsection{Monolingual Low-resource scenario}
The results of low-resource scenario are shown in parts D and E of \Tab{res-low-resource}. We call this low-resource because in this setting we used very limited or even no monolingual data for Japanese and English. We experimented with different low-resource settings: zero-resource setting (part D) where we have no Japanese and English data, limited-resource setting (part E) where we have 1M monolingual Japanese and English data. 

In part D, we observed large improvements, a maximum of 8.5 B\caps{leu} score over the baseline setting (A1), on all fine-tuning settings over model without fine-tuning when using only Chinese monolingual data (D1). Using only French data also gives better results on almost all fine-tuning settings, but not as large as that of using only Chinese data. When combining Chinese and French data, they seemed to conflict with each other leading to slightly reduced scores. 

In part E of the table, when there are 1M Japanese and English monolingual sentences, combining them with 20M Chinese data also gives improvements up to 1.1 B\caps{leu} points over A1. Combining with French data only gives occasional improvements. In this setting too, combining Chinese and French data led to reduction in performance.

Although French and English share cognates and have similar grammar, we have not performed explicit script mapping like we did for Chinese to make it more similar to Japanese. Nevertheless we have no satisfactory explanation for this unexpected phenomenon.
French-English token-level mapping might benefit the model better, but we leave this for the future.

We can draw the following conclusions,

\noindent\textbf{1.} Script mapping (Chinese to Japanese) will consistently give better results.\\
\noindent\textbf{2.} Using a linguistically similar languages (French and English) will sometimes give better results.\\
\noindent\textbf{3.} There may be conflicts between data of different assisting languages.

\subsubsection{Monolingual resource-rich scenario}
In part F, we found that there is less need to combine related language data when we use a large monolingual data of target languages. Only combining with Chinese data (F2) is comparable with pure Japanese-English monolingual pre-training (F1). Combining French data degrades the translation quality in most settings. Thus, assisting languages become interfering languages in scenarios where large amounts of monolingual data are available for languages to be translated.

\subsubsection{Unrelated language VS related language}

In part G of the table, we compare pre-training on related languages versus unrelated languages. We saw that using Arabic and Russian as unrelated assisting languages gives about 0.1 to maximum 1.5 improvement over the baseline (A1). This is surprising and it shows that leveraging any additional language is better than not leveraging them. However, using Chinese and French yields about 2 to 2.7 B\caps{leu} score improvements. Although we have only performed simple mapping of Chinese to Japanese and none for French to English, consider this result with the results in part B of the table where we contrasted script mapping and without script mapping. This gives a clear answer that increasing relatedness on top of already using related languages is definitely important. In the future, we will consider more rigorous ways of increasing relatedness between pre-training corpora by using existing dictionaries instead of simple script mapping.

\section{Conclusion}
In this paper we showed that it is possible to leverage monolingual corpora of other languages to pre-train NMT models for language pairs that lack parallel as well as monolingual data. We showed that maximizing the similarity between the other (assisting) languages and the languages to be translated is crucial and standard script mapping helps improve the translation quality.  We further showed that simple data selection methods help make the pre-training corpora more similar to the fine-tuning corpora which positively impacts NMT performance. In the future, we plan to experiment with even more challenging language pairs such as Japanese--Russian and attempt to leverage monolingual corpora belonging to diverse language families.We might be able to identify subtle relationships among languages and approaches to better leverage assisting languages for several NLP tasks.

\bibliographystyle{named}
\bibliography{reference}

\end{document}